# Machine Learning Techniques to Identify Hand Gestures amidst Forearm Muscle Signals

**Ryan Cho[a], Sunil Patel[b], Kyu Taek Cho[c], and Jaejin Hwang[b*]**

[a]Illinois Mathematics and Science Academy, Kane, IL, USA

[b]Department of Industrial and Systems Engineering, Northern Illinois University, DeKalb, IL, USA

[c]Department of Mechanical Engineering, Northern Illinois University, DeKalb, IL, USA

Corresponding author's Email: jhwang3@niu.edu

**Abstract**

This study explored the use of forearm EMG data to distinguish eight hand gestures. The Neural Network and Random Forest algorithms were tested on data from ten participants. The Neural Network achieved 97.13% accuracy on 1000-millisecond windows, while the Random Forest had 84.77% accuracy on 200-millisecond windows. Larger window sizes led to better gesture classification due to increased temporal resolution. The Random Forest was faster, with a speed of 91.82±10.34 milliseconds, compared to the Neural Network's 123.56±11.94 milliseconds. In conclusion, the study found that a Neural Network based on a 1000-millisecond stream was the most accurate, with an accuracy rate of 97.13%, while a Random Forest based on a 200-millisecond stream was the most efficient, with an accuracy rate of 84.77%. Future studies should increase the sample size, include more hand gestures, use different feature extraction methods, and modeling algorithms to improve the accuracy and efficiency of the system.

## 1. Introduction

Extended Reality (XR) technologies such as Virtual Reality, Mixed Reality, and Augmented Reality use devices such as smartphones and embedded displays to provide an immersive experience between virtual and physical realities. This technology has seen exponential development in demand in various industries, including entertainment and healthcare applications, in recent years. This rise in demand may be seen in the rising usage of XR in industries including healthcare and entertainment.

XR advances have certainly transformed several industries, but they are not without challenges. Existing computer-vision-based hand tracking systems for XR applications feature cause discomfort, which can lead to problems such as gorilla arm syndrome and rotator cuff injuries. Many studies have been conducted in the past to examine the effects of biomechanical exposure on the neck and shoulder from XR applications, with results indicating a high susceptibility to musculoskeletal pain and injury risks in XR applications [1, 2]. To avoid serious injuries, it is recommended that the vertical target site be 15° below eye height [3]. However, because of the very short range, this basic guideline is unlikely to be followed.

Another major problem is that computer vision based XR applications introduced use inadequate hand categorization methods. To capture every physical movement, these systems require a high frame rate, with each acquired frame transmitted to classification techniques [4]. Each captured frame is represented as a high-dimensional picture, resulting in more pixels and increased processing requirements during categorization. Due to the significant processing resources required to classify gesture movement, it is typical for XR settings to display very unstable virtual realities [5]. Additionally, computer vision systems perceive body motions ineffectively due to potentially complex backdrops linked with lighting conditions and low-resolution cameras that give inaccurate categorization conclusions [6].

Electromyography (EMG) sensors, which monitor electrical activity in muscle cells, have lately gained popularity as a means of addressing the aforementioned issue. Using EMG sensors instead of a computer vision-based hand tracking system resulted in more successful gesture classification with lower computing requirements. Instead of requiring high frame rates and processing on highly dimensional picture data, EMG sensor data can be divided into short clips with very few features for classification. In comparison, phone cameras are typically 640 by 320 pixels with 3 dimensions for color, implying that computer vision-based hand tracking systems for XR applications employ 614400 characteristics for each image categorization. More crucially, the location and lighting of a user's hand are irrelevant to EMG sensors because they always actively record the muscular activity generated on the hand. Due to the expensive cost of EMG sensors, they were formerly used mostly in medical equipment rather than XR applications. Nonetheless, the development of low-cost and high-quality EMG sensors is regaining attraction in XR applications.

Prior to conducting this research, a comprehensive review of existing EMG studies on hand gesture classification was created [7]. One such study utilized a wavelet decomposition approach with the Daubechies wavelet and extracted 12 conventional features along with 5 new features from the signal. The study aimed to classify 10 distinct hand movements using a neural network and compared model accuracy across varying signal lengths. The findings indicated an impressive algorithm accuracy of 95.5% for a signal length of 800 milliseconds, with shorter signal lengths indicating lower accuracy levels and longer signal lengths having higher accuracies. Due to these imbalances in accuracy in shorter signal lengths, the paper finalized its algorithm at an 800 milliseconds moving window size. However, it is worth noting that this may result in a potential issue with delayed execution time. Additionally, there are other factors that may impact the study's findings, such as the duration of time it takes to process each classification. If the duration of time required to classify each EMG signal is extended to lengths such as 800 milliseconds, it would pose problems for time-sensitive activities, particularly for assistive technologies used in tasks like lifting tree logs using cranes. A previous study found that delays in joystick response times exceeding 500 milliseconds during virtual reality experiences led to a noticeable decrease in the number of logs that could be successfully loaded [8].

The objective of this research was to investigate the performance of distinguishing various hand gestures using the EMG data based on two widely used machine learning algorithms, neural network, and random forest. This comparison will be based on several factors including conventional features, computing speed, algorithm accuracy, and the response to spontaneous reactions from various signal moving windows. This EMG study has features from multidomain analysis to improve the perception of the EMG signals from different domains of analysis. Time domain analysis will allow for understanding of the signal’s amplitude over time and the frequency domain analysis will allow for understanding the power spectral densities of different oscillations. Combining multiple domains of analysis have commonly been used for EMGs to increase algorithm accuracies [9]. Following that, various machine learning models and their processing expenses at different signal lengths were investigated to determine which approach is the best. It was hypothesized that even with small amounts of features and simple machine learning models, great levels of accuracy and simpler computation can be achieved in XR-based environments.

## 2. Methods

### *2.1 Participants*

Ten university students (7 male and 3 female) volunteered to be recorded using EMG sensors during various hand gesture actions. Their height in centimeters, weight in kilograms, and age in years have the following means and standard deviations: 169.0 ± 8.8 cm, 66.6 ± 14.4 kg, and 23.8 ± 1.3 years. Before beginning the data gathering process, each of these individuals agreed to sign a written consent form from the Institutional Review Board (IRB).

### *2.2 Experimental Protocol*

To measure muscle activity during extended reality (XR) hand gestures, two Delsys Trigno electromyography (EMG) sensors (Delsys Inc., Boston, MA) were placed on the flexor digitorum superficialis (FDS) and extensor digitorum communis (EDC) muscles. Before placing the EMG sensors on the muscles, the skin’s hairs were removed if needed and cleaned with alcohol to reduce the noise of the signal according to the EMG standard procedures [10]. Participants were asked to perform each of the eight hand gestures commonly used in XR applications. They were instructed to follow the order of activities as seen in Figure 1 and repeat each gesture ten times while wearing the EMG sensors. The EMG sensors recorded muscle activity in microvolts at a sampling rate of 1000 Hz, which was the default sampling rate offered by the Delsys Trigno EMG sensors. The data collected was based on hand gesture movements that were performed relatively slowly, meaning that 1000 Hz was suitable to capture this information. Participants were asked to perform each of the eight hand gestures commonly used in XR applications. They were instructed to follow the order of activities as seen in Figure 1 and repeat each gesture ten times while wearing the EMG sensors. Each hand gesture movement containing 10 repetitive cycles were recorded as a continuous time series. Once the hand gesture movement data collection was finished, they were saved into a comma-separated value file to use later for analysis.

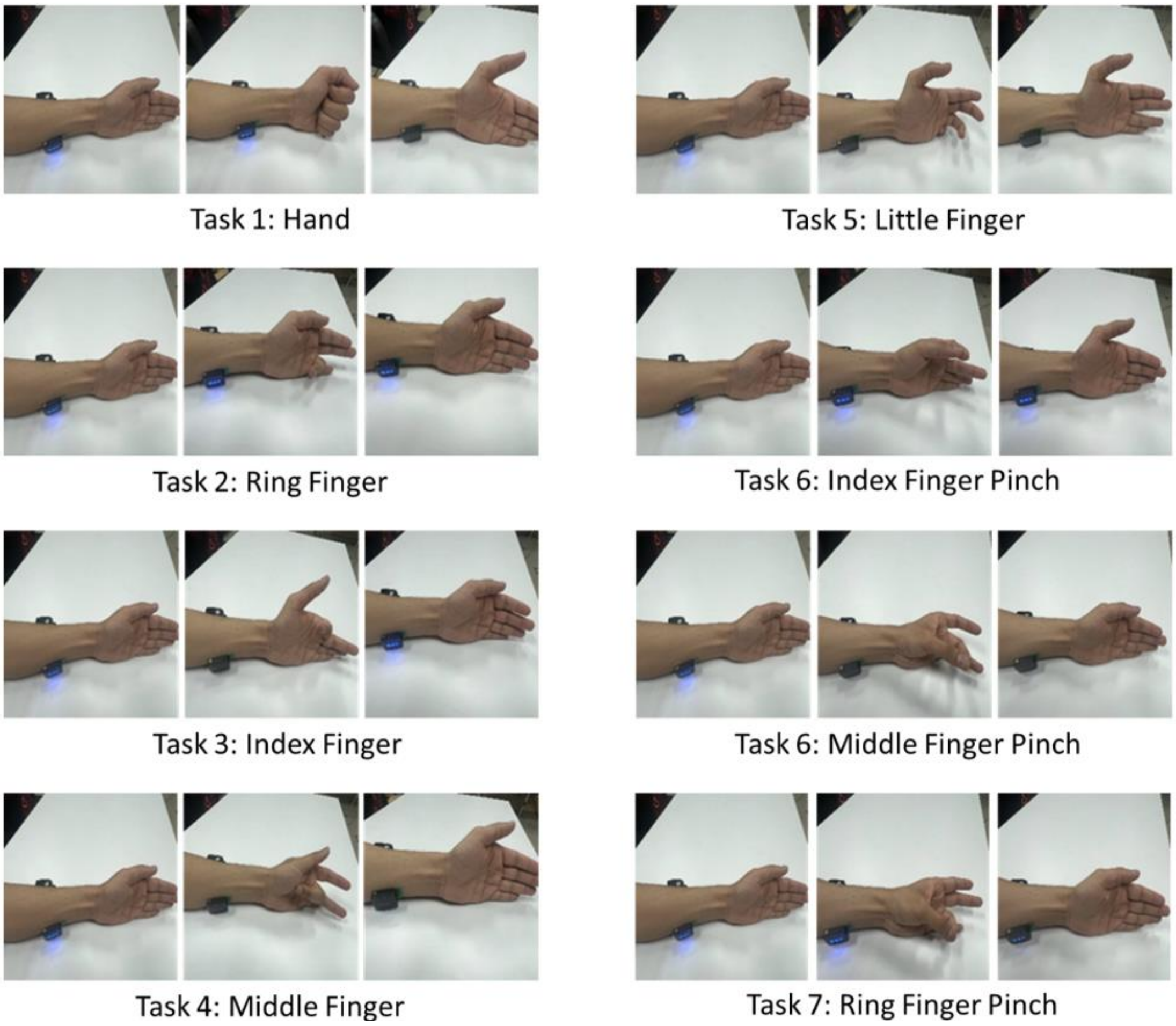

Figure 1. Various hand gestures tested in this study.

### *2.3 Signal Preprocessing*

EMG signals detect electrical activity in specific regions of the body. However, they are prone to picking up unrelated signals such as heartbeats, noise from electronic sensors, and signal instabilities. To address these issues, most literature employs frequency-selective approaches such as bandpass filters and resampling. Unfortunately, these approaches greatly reduce the original signal data when attempting to remove noise [11]. Instead, wavelet denoising approaches that use the time and frequency domains simultaneously should be used for denoising, as they can retain more signal data than frequency-selective methods based on the Fourier Transform that only employ information in the frequency domain.

In previous studies, the Daubechies order 6 wavelet with 4 layers of decomposition has shown very high levels of signal elimination, making it a suitable approach for EMG data denoising [12]. This wavelet features spiky patterns that match

EMG patterns, resulting in greater signal retention during signal reconstruction stages following denoising. Together with this wavelet, a soft Bayes Shrink threshold was employed, which is a typical approach for image denoising but can also be used for one-dimensional signals like EMGs.

During each level of wavelet denoising, the signal was divided into two equal subbands based on frequency: approximate coefficients and detail coefficients. These subbands represent successive stages of the wavelet decomposition. In this study, the sampling rate was 1000 Hz, implying that the highest frequency expressed, as per the Nyquist-Shannon Sampling Theorem, is 500 Hz. Consequently, during the first level of decomposition, the approximate coefficient includes 0-250 Hz time series data and the detail coefficient includes 250-500 Hz time series data. Each subband is assigned a distinct threshold determined by a Generalized Gaussian Distribution (GGD). After applying these thresholds to the wavelet coefficients, an inverse wavelet transform is performed to obtain the denoised estimate [13]. As the denoising process proceeds to subsequent levels of decomposition, the approximate coefficient subband from the first level is once again split into the two subbands (approximate coefficient and detail coefficient) as mentioned earlier. This process is repeated for every denoising level.

To improve training accuracy, the denoised data was segmented into clips with a designated clip size, with 90% overlap between adjacent windows. Overlapping windows in signals help retain commonly found signal characteristics through larger temporal resolution ranges, improving algorithm memory retention [14].

### *2.4 Time Domain Features*

A commonly used method to classify these gestures is using the time domain features. These features are known to be incredibly efficient due to their computational simplicity and representing the transitory response of bio-signals. The features shown in Table 1 were used during extraction.

Table 1. Time domain features commonly used: with x representing a list of all the time series data points of the signal and n representing each index of the list sequentially.

| **Features** | **Equations** |
|---|---|
| Integrated EMG (IEMG) | $\sum_{n=1}^{N} \lvert x[n] \rvert$ |
| Integrated Absolute of Second Derivative (IASD) | $\sum_{n=1}^{N-2} \lvert x'[n+1] - x'[n] \rvert$ |
| Integrated Absolute of Third Derivative (IATD) | $\sum_{n=1}^{N-3} \lvert x''[n+1] - x''[n] \rvert$ |
| Integrated Exponential of Absolute Values (IEAV) | $\sum_{n=1}^{N} \exp(\lvert x[n] \rvert)$ |
| Integrated Exponential (IE) | $\sum_{n=1}^{N} \exp\,(x[n])$ |
| Mean Absolute Value (MAV) | $\frac{1}{n}\sum_{n=1}^{N} \lvert x[n] \rvert$ |
| Root Mean Square (RMS) | $\sqrt{\frac{1}{n}\sum_{n=1}^{N} \lvert x[n] \rvert}$ |
| Variance (VAR) | $\frac{1}{n-1}\sum_{n=1}^{N} x[n]^2$ |
| Zero Crossing (ZC) | $\frac{1}{2}\sum \lvert x[n] - x[n-1] \rvert$ |
| Waveform length (WL) | $\sum_{n=1}^{N-1} \lvert x[n+1] - x[n] \rvert$ |

The Integrated Absolute of Second Derivative is a mathematical approach for calculating the rate of change of a signal. It detects certain properties such as signal concavity, inflection points, and other changes in the curvature of the signal and is beneficial in locating features that are less influenced by noise in the signal. This technique is widely employed in signal processing and data analysis applications [7].

The Integrated Absolute of Third Derivative is a mathematical technique for calculating the rate of change of a signal's rate of change. This method is successful in filtering away noise from the signal and is used to capture certain aspects such as rapid changes in the signal's curvature. This approach is frequently utilized in signal processing applications requiring great precision [7].

The Integrated Exponential of Absolute Values is a mathematical technique for amplifying bigger amounts of data while suppressing smaller amounts. This method is used to detect crucial elements in a signal that would be difficult to detect using other techniques. It is frequently employed in applications where detecting minute but substantial changes in a signal, such as medical or scientific research [7].

The Integrated EMG function amplifies positive samples while suppressing negative sections of a signal. It is often used to evaluate muscle activation in electromyography (EMG) studies. This method can be utilized in rehabilitation or sports training applications to identify and measure the intensity of muscular contractions [7].

The Mean Absolute Values approach is a mathematical technique for calculating a signal's average absolute amplitude. It offers information about the general amplitude of the signal and can be used to determine trends or variations in the signal over time. This technique is widely utilized in data analysis and signal processing applications [15].

To calculate the total energy of a signal, mathematicians utilize the Root Mean Square approach. In applications where quantifying the signal's strength is crucial, like in audio or vibration analysis, it gives fundamental information about the signal's power or intensity [15]. A statistical tool for gauging how much a signal changes over time is the variance method. It helps to spot patterns or trends in the signal by revealing information about the signal's stability or variability. Applications for machine learning and data analysis frequently use this technique [16].

A method called the Zero Crosses approach can be used to determine a signal's frequency. It helps discover periodicity or oscillations in the signal by counting the instances when the signal crosses the y axis. Applications for audio and communication frequently employ this technique [7].

A mathematical method for calculating a signal's total length is the Waveform Length method. It is helpful in locating changes or abnormalities in the signal and offers information about the level of activity in the signal. During extraction, this technique is frequently employed in signal processing and data analysis applications [5].

### *2.5 Frequency Domain Features*

Frequency domain features are extracted by transforming time domain signals into power spectral density (PSD) over different frequencies. This transformation yields various types of features that can be analyzed to extract useful information. The commonly used approach for creating frequency domains is through Fast Fourier transform (FFT), which decomposes the signal into different frequency components. However, this method has high variance levels, leading to biased data, and requires stationary signals, which may not be applicable to EMG signals based on XR applications that change quickly over time. Two other methods that can be employed are the Multitaper and Welch Periodogram. The Multitaper Periodogram uses multiple Slepian tapers that provide independent estimates of spectral content at different locations on the EMG data. The FFT is applied to each of these tapered time series to convert them into the frequency domain, resulting in a series of Tapers that are then averaged to create a Multitaper Periodogram. Although this method is computationally intensive, it reduces variance while maintaining high frequency resolution [17].

The Welch Periodogram, in contrast, divides an EMG signal into several small windows of data, and each window is subjected to a FFT before averaging to yield the final result [18]. While this approach reduces variance, it also lowers frequency resolution, and the degree of reduction is proportional to the number of window cuts in each EMG signal, as depicted in Equation 1.

$$F_{res} = \frac{F_s}{N} = \frac{F_s}{F_s t} = \frac{1}{t} \tag{1}$$

where, Fs denotes the original signal sampling frequency, N represents the number of windows being used in the periodogram which in this case is 2, and t is the total time duration of the signal.

Although both the Multitaper and Welch Periodogram have limitations, the latter was chosen for several reasons. Firstly, in XR applications, computation is a crucial factor, making the Multitaper Periodogram less applicable to this study due to its higher computational cost. Moreover, empirical evidence suggests that both methods perform similarly, except when signals are particularly noisy, which can decrease precision in the Welch Periodogram, while the Multitaper Periodogram remains unaffected. However, this issue can be mitigated by ensuring correct sensor connections and using appropriate data normalization techniques. To maintain adequate frequency resolution, a window size equal to 90% of the EMG signal clip for each FFT application with 80% clip overlaps throughout the signal was used to compute the Welch Periodogram. The Welch Periodogram was then applied to extract features presented in Table 2.

Table 2. Commonly used frequency domain features: P represents a list of all the data points inside the create periodogram, f represents a list of frequencies used in the x axis of the periodogram, j represents each index number being used in the periodogram, and M represents the maximum frequency limit, which in this case is 500 Hz.

| **Features** | **Equation** |
|---|---|
| Mean frequency (MF) | $\sum_{j=1}^{M} \frac{f_j P_j}{P_j}$ |
| Median frequency (MDF) | $\frac{1}{2} \sum_{j=1}^{M} Pj$ |
| Peak frequency (PF) | $argmax_f P(f)$ |
| Spectral Entropy (SE) | $\sum_{i=1}^{N} P_i(f) \log_2(P_i(f))$ |
| Total power (TP) | $\sum_{j=1}^{M} P_j$ |
| Mean Power Frequency (MPF) | $\frac{\sum(f \cdot P(f))}{\sum P(f)}$ |
| Signal to Noise Ratio (SNR) | $\sqrt{\frac{\sum_i^n (s_i - \bar{s})}{n}}$ |
| Spectral Edge Frequency (SEF) | $SEF = \int_{0.5 f_{max}}^{f_{\max}} P(f) df$ |
| Frequency Ratio (FR) | $\frac{\sum_{j=LLC}^{ULC} (P_j)}{\sum_{j=LHC}^{UHC} (P_j)}$ |

The average frequency of a signal is derived as a typical metric in signal processing and is weighted by the power at each frequency bin. It offers details on the signal's frequency makeup and can be used to pinpoint the main frequency bands within the signal. [19]

Another statistic used to assess frequency content is the median frequency, which is the frequency at which half of the signal's strength is below and half is above. It can be used to identify variations in a signal's frequency distribution and provides details about the signal's center frequency [14].

The frequency at which the signal is most powerful is known as the peak frequency. It can be used to track variations in frequency content over time and to determine the dominant frequency in a signal [14].

Spectral entropy is a metric for signal complexity that shows how evenly the energy of a signal is dispersed across its frequency spectrum. When noise or other disturbances are present, it can be utilized to detect changes in the complexity of the signal [20].

Total Power is the sum of the power at each frequency bin and represents the total energy in a signal. It gives details about a signal's overall strength and can be used to contrast transmissions with various frequency contents [21].

The Mean Power Frequency measures the average frequency where the power spectral density is distributed. This commonly used EMG feature has been used in the past to assess fatigue [22].

The Signal to Noise Ratio (SNR) is a metric used to compare the strength of the signal to the assumed noise range. In this investigation, it was assumed that noise made up the lowest 10% of the maximum power. SNR is frequently used to evaluate a signal's quality and has the ability to detect the presence of noise or other interference [23].

The frequency below which a particular percentage of the power of the signal is confined is known as the spectral edge frequency. In this study, a 95% confidence interval was chosen to emphasize this data. It can be used to detect changes in the frequency content of the signal and offers details on the distribution of energy throughout the frequency spectrum [24].

The power in the frequency ranges of 20–250 Hz and 250–1000 Hz are contrasted using the Frequency Ratio measure. It can be used to detect variations in frequency content over time and gives details about the relative strength of various frequency bands in a signal [25].

### *2.6 Machine Learning Models*

The features extracted from the data clips were first randomized by randomly mixing all the subject data and their corresponding gestures. Following this randomization step, 80% of the data was designated for algorithm training, while the remaining 20% was allocated for testing. This approach allows the algorithms to be trained using data from the entire group and assessed based on their performance with previously unseen data. However, before adding in the extracted features into the machine learning algorithms, a Standard Scalar method was applied onto the data. This Standard Scalar can help even out abnormally high pieces of data to avoid uneven classification results in machine learning algorithms. Two different algorithms were evaluated: Random Forest and a Neural Network.

These two algorithms were specifically chosen due to their popularity and distinct differences during classification processes. Random Forest utilizes the collective predictions of multiple algorithms and identifies important features, giving it the advantage of ranking features fed into the algorithm based on importance. This significantly helps during classification. On the other hand, Neural Networks leverage interconnected layers of neurons to uncover intricate patterns for classification, making them particularly good at recognizing complex patterns. In previous studies, both of these algorithms have proven to be very accurate classifiers compared to other methods. For instance, a previous study, which compared EMG signals during 10 different hand motions using the neural network, support vector machine, logistic regression, and random forest, it was found that the neural network performed the best with an accuracy of 94% [26]. Similarly, another study comparing the support vector machine, naïve Bayes, random forest, and k-nearest neighbor, found that the random forest performed well at a 98% classification accuracy [27]. Overall, this comparison between these two popular and distinct methods of classification allows this study to focus on only the most relevant algorithms.

Random Forest is a unique method that creates multiple decision trees from subsets of observations in the dataset. These trees are then aggregated to find the majority voting for the classification result. To achieve an optimal balance between underfitting and overfitting the data, we employed a hyperparameter optimization system based on 5-fold cross-validation to determine the ideal number of decision trees for our algorithm. This systematic approach ensures that we avoid the pitfalls of using too few or too many trees. The effectiveness of this technique has been demonstrated in a previous study that focused on classifying hand movements from EMG signals using the random forest algorithm, achieving an impressive classification accuracy of 95.39% [28]. However, unlike the compared study, which used 9-fold cross-validation, our study opted for 5 folds. The choice of 5 folds allowed for faster training, which was essential due to the extensive list of moving window sizes being tested. Despite using fewer folds, this approach still provides an effective means of evaluating the presence of underfitting or overfitting based on the number of decision trees present. By finding the best number of trees for the algorithm based on 5 k fold cross validation, the Random Forest can achieve the highest possible accuracy.

The neural network consists of layers of interconnected nodes that transmit signals between each other. Each node has a unique weighting that alters signals transmitted through different nodes. Using 60 nodes in the first layer was used even though this algorithm only takes in 18 features from each EMG signal, totaling up to 36, can allow the algorithm to understand the features being fed into it more easily as there is more nodes to use for understanding these patterns. Then, the use of three hidden layers with 1000 nodes each is that it helps the neural network to fully capture complex patterns and relationships within the data, enabling it to learn more intricate features through the data. The final layer used 8 nodes since after all, the algorithm is classifying between 8 different hand gestures. The hidden layers were all applied with the ReLU activation function to help the model handle non-linearities and the sigmoid activation function allows the algorithm to create predictions on which output is most likely the classification result. Algorithms in past literature have also used rather complex neural networks for EMG hand classification, such as Azhiri et. al whom used six hidden layers in their neural network [7]. However, this study decided to modify using six hidden layers to just three to avoid super long computational runtimes, especially since the neural network uses a more complex algorithm structure compared to the random forest. The sigmoid activation on the final layer of the neural network allows the function to predict the probability of the prediction result, ranging between 0 and 1. A value closer to 1 indicates higher confidence in the predicted outcome. In this specific scenario, the neural network is trained to classify eight different hand gestures. The sigmoid activation function on the final layer generates probabilities for each gesture being the classification result. Each neuron in the final layer corresponds to one of the eight gestures, outputting the probability of that gesture being present in the input data. During the training of the neural network, a batch size of 64 was used for better algorithm generalization and to lower the computational expenses when training the algorithm. The algorithm was compiled with an Adam optimizer and categorical cross entropy loss function. The Adam optimizer allows the algorithm to adapt its learning rate during training based on the gradients of the loss function. The loss function was based on categorical cross entropy which calculates the dissimilarity between the algorithm predictions and the actual results, allowing the algorithm to learn from its mistakes. The list of features were designed and extracted based on the previous study investigating the EMG-based hand gesture classification using the neutral network algorithms [7].

Evaluation metrics such as F1 score, precision, recall, and accuracy were used to determine the performance of the machine learning model. The F1, precision, and recall score helps to identify signs of algorithm bias by considering if a particular classification label outnumbers others. Although the dataset had relatively equal amounts of each classification label in both the training and testing dataset, these evaluation metrics can help to identify specific mistakes made by the algorithm using True Positives, False Positives, and False Negatives, as seen from Equation 2. The overall test data accuracy was also recorded by calculating the ratio between the number of correctly classified samples and the total number of samples.

$$\text{Precision:} \quad \frac{True\ Positive}{True\ Positive + False\ Positive} \tag{2-a}$$

$$\text{Recall:} \quad \frac{True\ Positive}{True\ Positive + False\ Negative} \tag{2-b}$$

$$\text{F1 score:} \quad \frac{True\ Positive}{True\ Positive + \frac{1}{2}(False\ Positive + False\ Negative)} \tag{2-c}$$

While the noise removal method, feature extraction from both time and frequency domains, and machine learning algorithms are undeniably useful, their computational demands can impede the performance of users in XR applications. A previous study found that for remote assistive technology integrated with XR, which can involve time sensitive activities, maintaining joystick delays under 500 milliseconds was crucial for enabling users to excel in these XR applications [8]. To assess their computational costs, the processing speed of each step, including noise removal, feature extraction, and machine learning classification, was measured, and analysed to determine the best methods to use or whether a particular method should still be incorporated into the pipeline.

# 3. Results

## *3.1 Noise Removal Effects*

Prior to evaluating the efficiency, we employed the wavelet denoising technique, utilizing the Daubechies order 6 wavelet with 4 levels of decomposition and the soft Bayes Shrink thresholding method. To illustrate this denoising approach, we utilized the EMG signal obtained from data collection subject 3 during hand motion number 3 from one trial repetition, covering a complete cycle of the hand gesture. Figure 2-(a) displays the signal before denoising, whereas Figure 2-(b) showcases the signal after denoising.

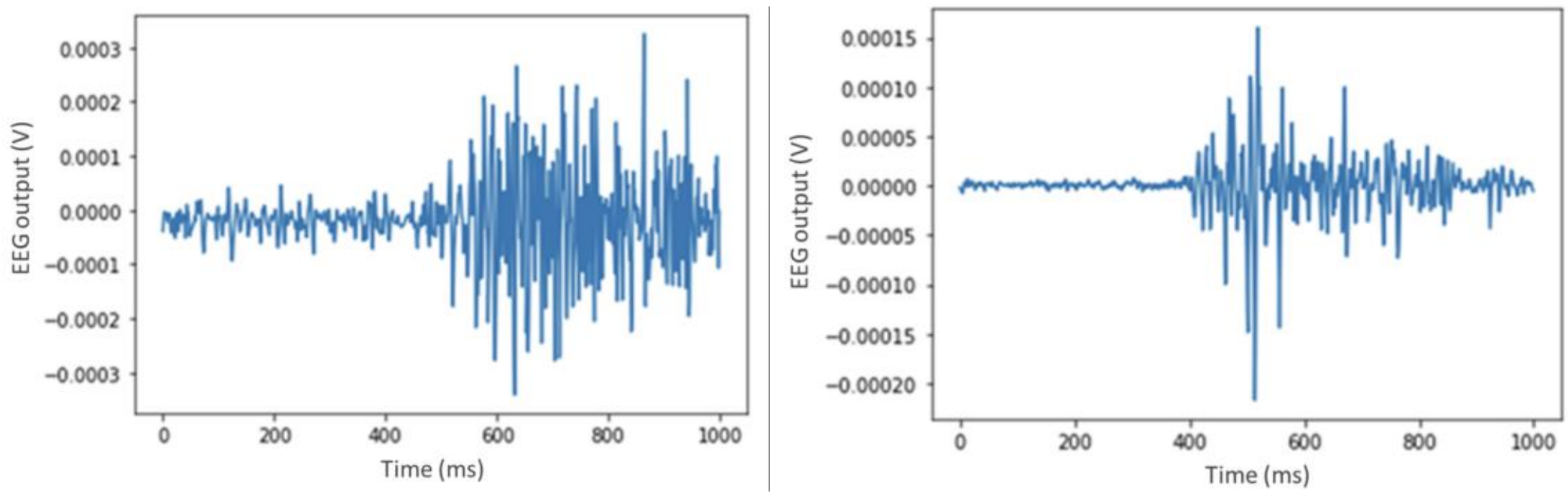

(a) (b)

Figure 2. One data sample from subject 3 performing hand gesture 3: a) raw signal, b) wavelet denoised signal.

To statistically measure the amount of noise removed, we employed the signal-to-noise ratio calculation, as illustrated in Table 1, for each signal from various moving window sizes both before and after wavelet denoising. The results in Table 3 demonstrate that after applying wavelet denoising, the signal-to-noise ratio becomes higher than that of the raw data, indicating a significant increase in the signal's strength relative to the noise.

Table 3. Signal-to-noise ratios before and after wavelet denoising for various window sizes.

| Window Sizes (MS) | Raw SNR mean | Denoised SNR mean |
|---|---|---|
| 200 | 1.37 | 1.46 |
| 400 | 1.36 | 1.46 |
| 600 | 1.37 | 1.47 |
| 800 | 1.38 | 1.47 |
| 1000 | 1.38 | 1.47 |

### *3.2 Efficiency*

To evaluate the efficiency of different pipeline stages such as denoising, feature extraction, and machine learning classification, 12.7 GB of RAM was used to measure their execution times. This comparison provides a realistic assessment of computational runtimes in phone environments, as most phones typically have between 12 to 16 GB of RAM. The computational runtimes of each wavelet denoising technique were recorded for different window lengths, from which the mean and standard deviation were calculated. As illustrated in Figure 3, the wavelet denoising method was very fast, meaning that the application of wavelet denoising into the pipeline was not computationally costly while providing create noise removal methods at the same time. Although processing times may vary depending on device performance, all window sizes demonstrated comparable minute processing times, demonstrating that the computational expenses of denoising signals are not prohibitively expensive.

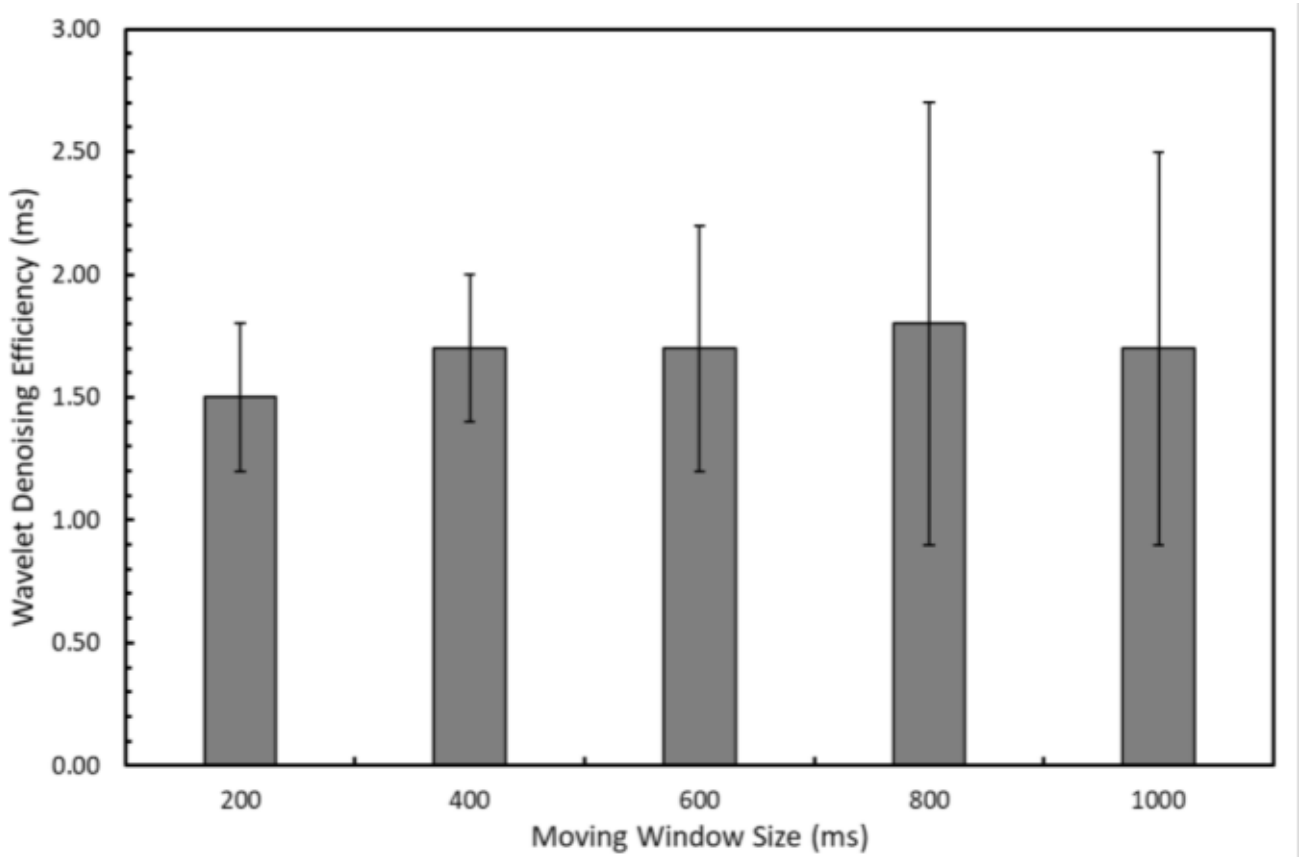

Figure 3. Mean (bars) and standard deviation (error bars) of denoising efficiency by different window lengths.

To assess the computational efficiency of features, the mean and standard deviation of the processing time required for each extraction method on different window sizes were measured. Specifically, the time it took to extract features from one piece of data for varying window sizes using the program was measured. The results, as shown in Figure 4, indicate that all features took approximately the same amount of time and prove to not be a problem computationally. The 200-millisecond moving window took 6.2 ± 1.3 milliseconds, and the 1000-millisecond moving window took 7.6 ± 2.4 milliseconds.

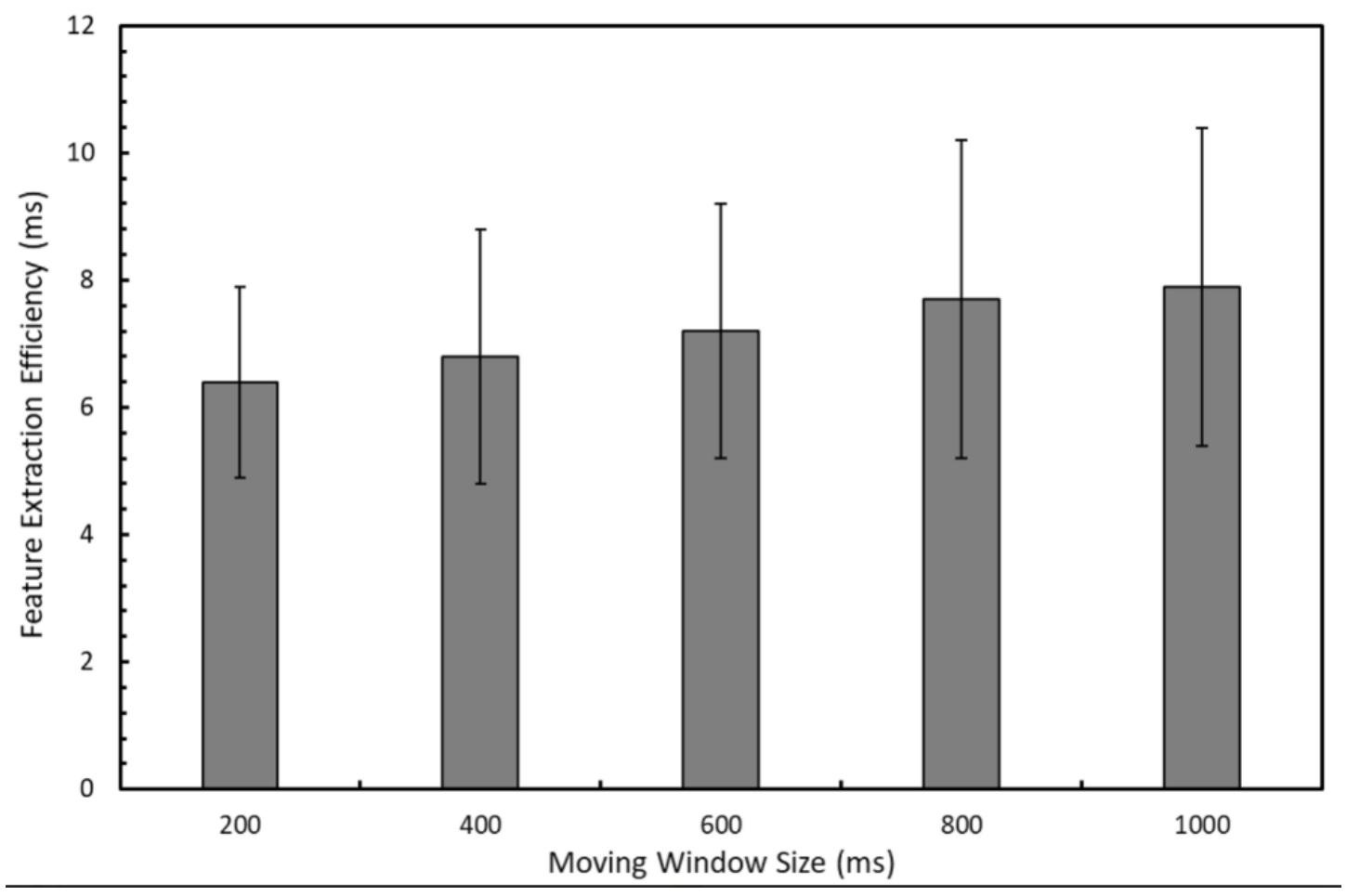

Figure 4. Mean (bars) and standard deviation (error bars) of feature extraction efficiency by different window lengths.

After determining that denoising and feature extraction steps do not significantly hinder the hand gesture differentiation pipeline, the next step was to assess the computational costs of machine learning classifications.

### *3.3 Machine Learning Model Results*

Subsequently, the testing dataset was used to evaluate two distinct machine learning models, as previously stated. To determine the performance of each algorithm, the study assessed several key metrics, including classification speed, testing accuracy, precision, recall, and F1 score.

Table 4. Comparison of algorithm performances.

| Algorithm | Accuracy | Precision | Recall | F1 score | Window |
|---|---|---|---|---|---|
| Random Forest | 84.77% | 84.83% | 84.78% | 84.76% | 0.2 s |
| Neural Network | 80.12% | 80.09% | 80.10% | 80.02% | 0.2 s |
| Random Forest | 91.07% | 91.14% | 91.07% | 91.07% | 0.4 s |
| Neural Network | 90.12% | 90.16% | 90.11% | 90.10% | 0.4 s |
| Random Forest | 93.84% | 93.90% | 93.84% | 93.83% | 0.6 s |
| Neural Network | 93.10% | 93.21% | 93.17% | 93.15% | 0.6 s |
| Random Forest | 95.12% | 95.13% | 95.12% | 95.10% | 0.8 s |
| Neural Network | 95.12% | 95.09% | 95.17% | 95.11% | 0.8 s |
| Random Forest | 96.65% | 96.67% | 96.65% | 96.66% | 1.0 s |
| Neural Network | 97.13% | 97.18% | 97.12% | 97.14% | 1.0 s |

The results presented in Table 4 indicate that the classification accuracy of the data segments is relatively consistent across window sizes and techniques. At a 200-milliseconds moving window size, Random Forest performed at an accuracy of 84.77 % and Neural Network had an accuracy of 80.12 %. When the moving window size increased to 1000-milliseconds, the Neural Network performed at 97.13 % and the Random Forest performed at 96.65 %.  Figure 5 includes the confusion matrix for the Random Forest's performance at 200-milliseconds and Figure 6 shows the confusion matrix of the Neural Network at 1000-milliseconds.

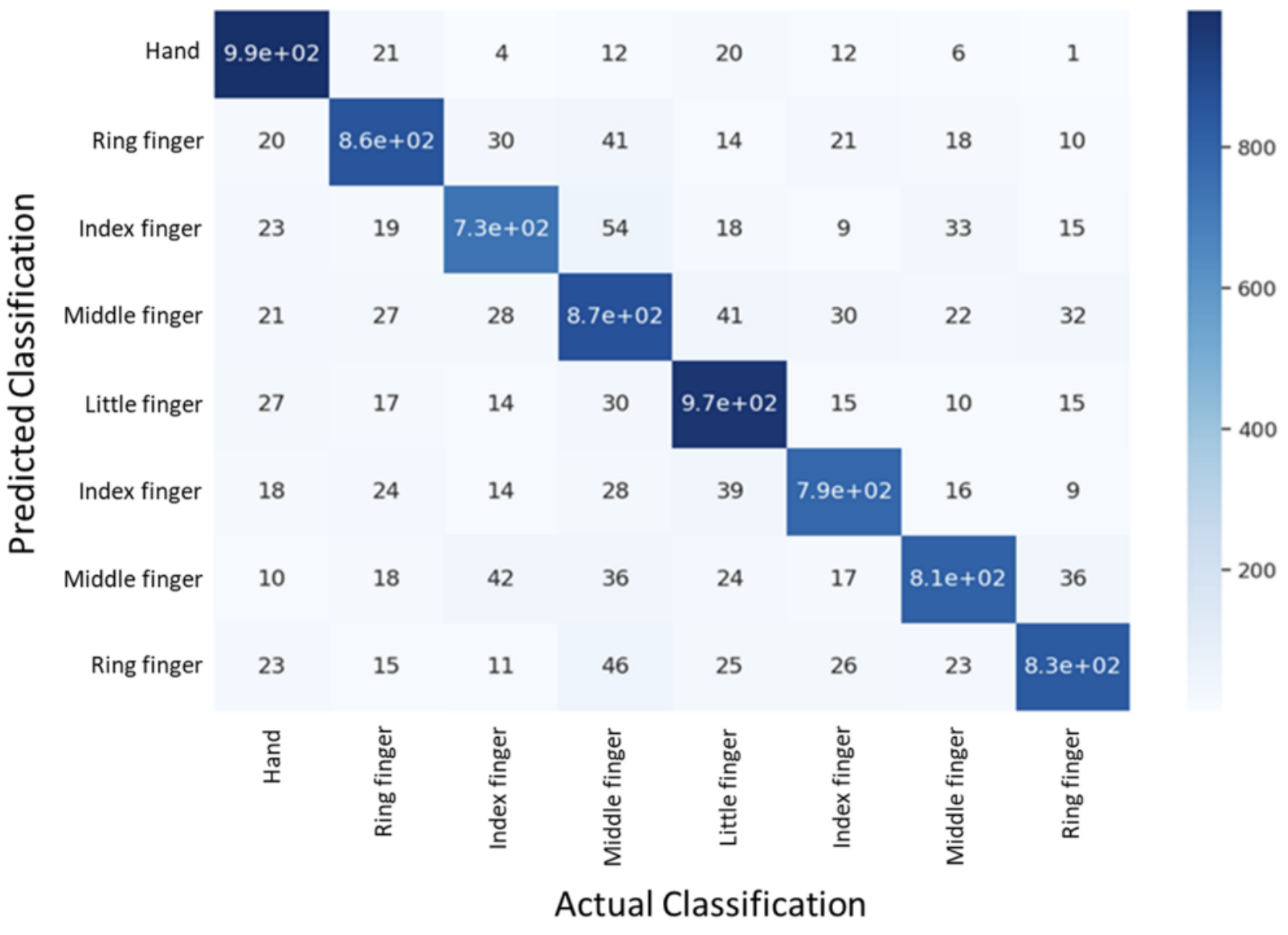

Figure 5. Confusion matrix of the random forest based on a 200-millisecond moving window.

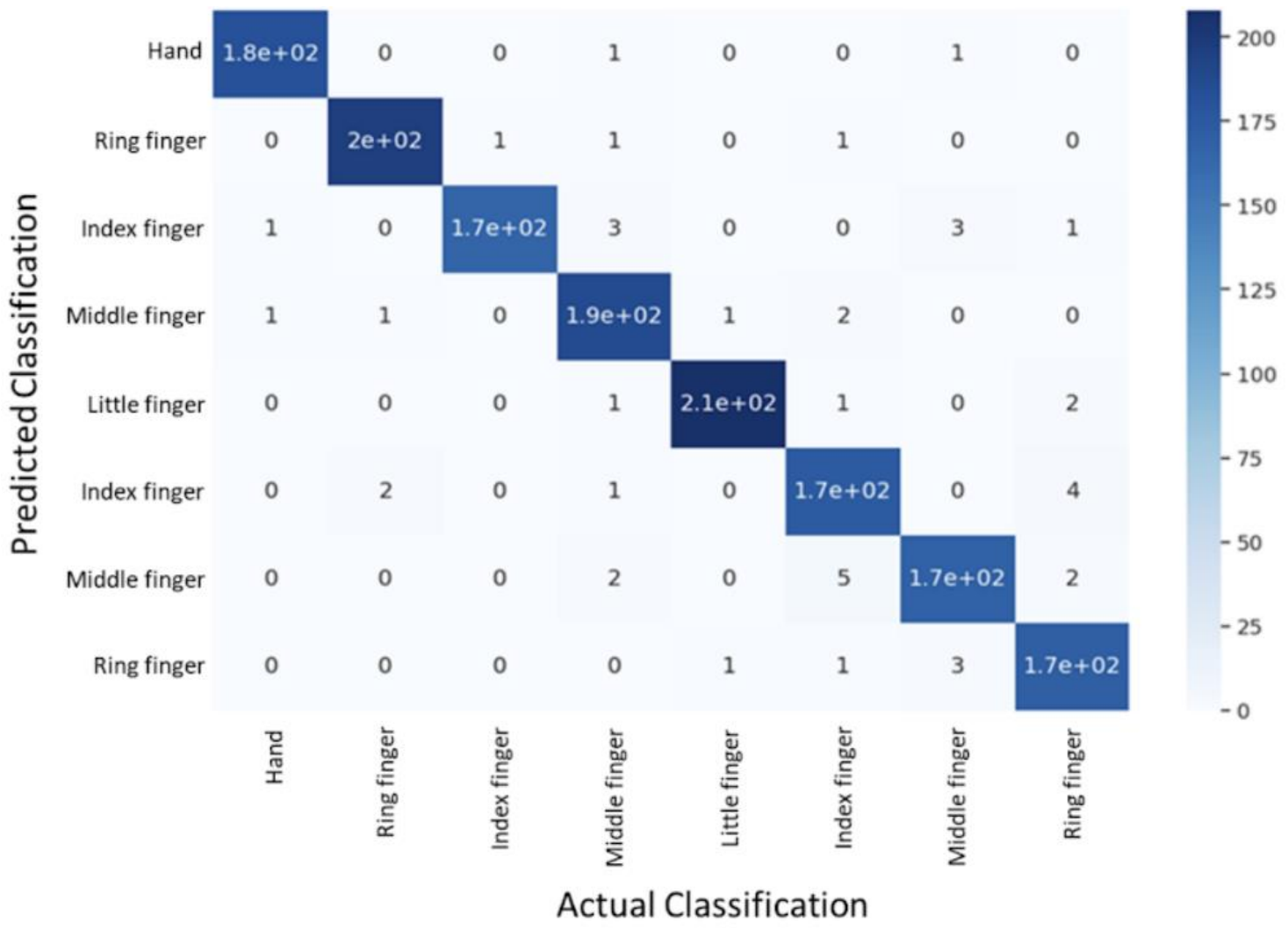

Figure 6. Confusion matrix of the neural network based on a 1000-millisecond moving window.

The algorithms were also tested for their computational runtime for each set of features pushed into them with the help of the Python "timeit" command-line interface, calculating the mean and standard deviation of the computational runtime. Figure 7 illustrates these results, with Random Forest performing at 91.82 ± 10.34 milliseconds and Neural Network performing at 123.56 ± 11.94 milliseconds.

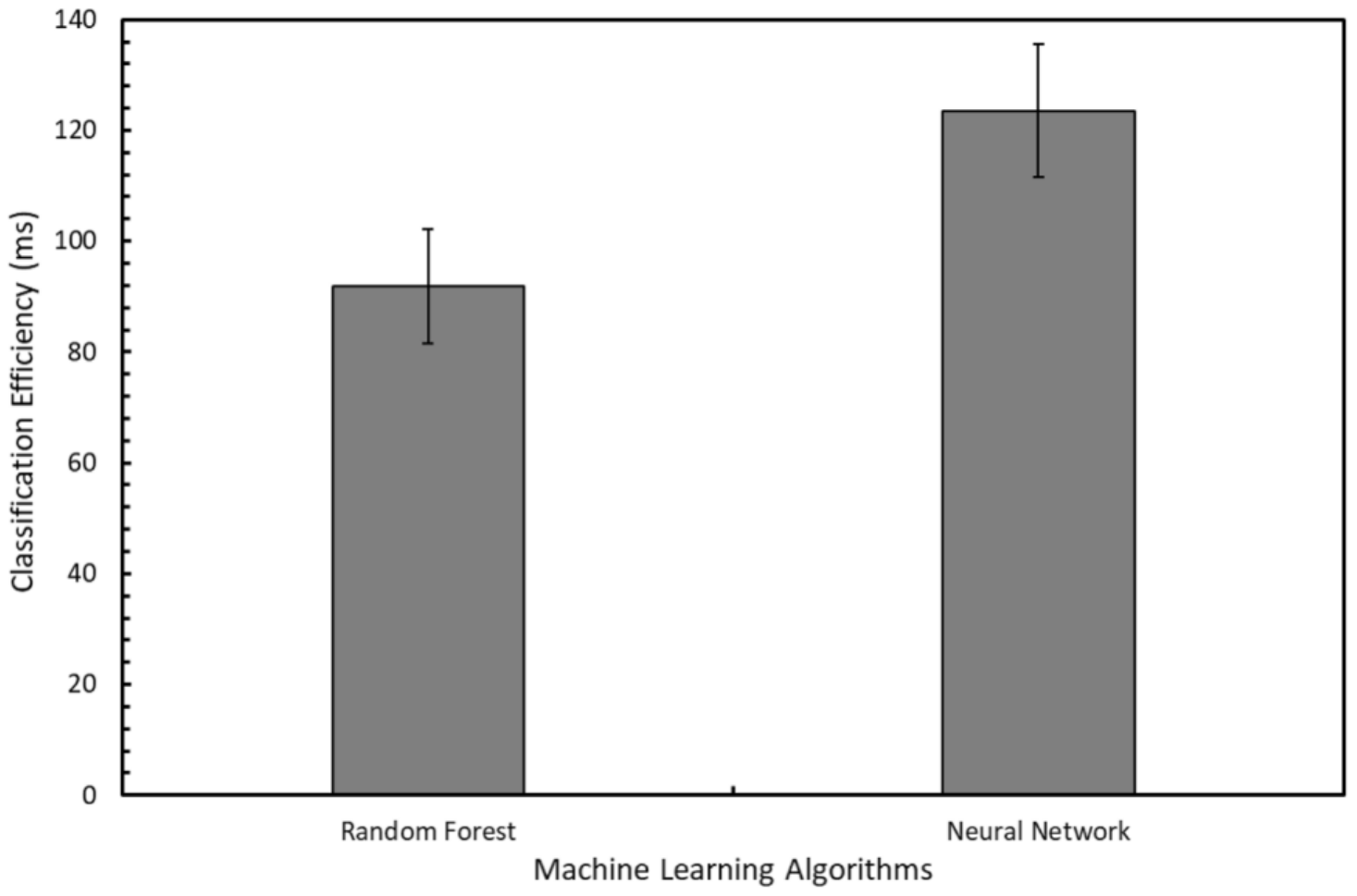

Figure 7. Mean (bars) and standard deviation (error bars) of algorithm classification efficiency.

## 4. Discussion

In this study, the aim was to evaluate classification methods for eight different hand movements commonly used in XR applications. Two EMG sensors were placed at various muscular locations and were analyzed for its tradeoffs between signal length duration and algorithm accuracy. The goal of the study was to enhance efficiency and reduce physical stress associated with XR application technology. Various feature extraction techniques across multiple domains of analysis, different window lengths, and machine learning algorithms were also tested to discover the best possible methods.

This study's findings showed that wavelet denoising is a crucial step. By retaining signal characteristics and removing noise, as seen from the stronger resulting signal-to-noise ratio seen in Table 3 and providing rapid computation by only taking 1.5 ± 0.3 milliseconds in the 200-millisecond moving window, wavelet denoising is a very helpful step in XR applications. The features extracted seen in Table 1 and Table 2 have also been found to be very quick, taking only 6.4 ± 1.5 milliseconds in the 200-millisecond moving window size. Additionally, it was discovered that the machine learning model classifications are remarkably quick, with no significant impact on feedback time. Notably, as seen in Table 4, the Random Forest algorithm performing on 200-millisecond intervals is very efficient, as it can apply a noise removal method in only 1.5 ± 0.3 milliseconds, extract features from the data in just 6.4 ± 1.5 milliseconds, and classifies the data in 91.82 ± 10.34 milliseconds. These results suggest that the proposed pipeline can perform accurate and rapid analysis of EMG data, making it a promising tool for a wide range of real-time applications.

It was also found that the accuracy metrics of the algorithms varied at different signal lengths in a distinct pattern. The shorter the signal length, the lower the algorithm accuracy became. This relationship between signal length and accuracy has also been seen by other literature [7]. From the University of Texas, researchers found accuracies in the 80 % range between 250 and 500 milliseconds, before the accuracy rose up and slowly plateaued at around 97 %. Because they chose an algorithm with a 95.5 % accuracy rate on an 800-millisecond signal length, they raised various concerns about the execution delay EMG machine learning systems provide. They chose the technique that produces data in 800 milliseconds because they thought accuracy was more important than signal efficiency.

While it may not seem like it, this interesting phenomenon of algorithm accuracy in relation to signal length creates an opportunity. Either XR applications utilize this study’s Random Forest algorithm at a 200-millisecond signal length, which emphasizes efficiency over accuracy with an accuracy rate of 84.77 %, or the Neural Network algorithm based on a 1000-millisecond moving window with an accuracy rate of around 97.13 %.

### *4.1 Real Life Applications*

EMG-based classification can be used in a variety of fields that require real-time and precise measurements. This approach can help improve the accuracy and efficiency of EMG-based XR applications, which can have significant implications for the development of assistive technologies for people with disabilities or injuries [29]. For example, it may be possible to create prosthetic limbs that can respond to the user's muscle movements in real-time, providing a more natural and intuitive interface. Regardless of what application, there should be a decision on whether efficiency of accuracy is more important, allowing people to either pick the Random Forest algorithm that performs very well on short pieces of data in 200 millisecond segments or the Neural Network algorithm that perform well on 1000 millisecond clips. Overall, this study highlights the potential of combining signal processing, feature extraction, and machine learning algorithms for EMG-based classification, and the importance of considering the specific requirements of a given application when choosing the appropriate technology.

### *4.2 Limitations and Future Direction*

The use of computer vision in Extended Reality (XR) applications offers several advantages, such as the ability to function without the need for additional external devices or sensors. However, the electromyography (EMG) approach used in this study used data based on several variations in muscle size and limb proportions. As a result, data collection using EMG technology may result in disparities in results between individuals with different muscular strengths and arm sizes, which may not be encountered by computer vision algorithms.

To overcome this limitation, this study aimed to gather a large and diverse dataset from individuals with varying muscular profiles to develop a generalized machine learning algorithm that can be applied to a wider population [30]. This approach could potentially enable a more inclusive and equitable use of XR technology in various fields, including assistive technologies for people with disabilities or injuries.

To enhance this study further, it is recommended to include a validation set during algorithm training. The absence of a validation set during training increases the risk of overfitting, resulting in reduced accuracy when dealing with new data. By incorporating a separate validation set, we can assess the algorithms' performance on unseen data, identify overfitting, and optimize their accuracy and robustness.

Despite its limitations, it is important to recognize that the EMG approach offers several potential benefits, such as providing more precise and accurate measurements in certain applications, including fine motor control, and detecting subtle changes in muscle activity. Therefore, it is crucial to carefully consider the specific requirements of a given application and choose the appropriate technology accordingly. In the future, EMG technology could be improved by reducing the size and cost of sensors, simplifying calibration processes, and developing more advanced algorithms to account for individual differences in muscular profiles.

## 5. Conclusions

In summary, this study has developed an optimal classification approach that addresses the tradeoff between accuracy and efficiency in EMG-based classification for XR applications. By utilizing noise reduction techniques and effective feature extraction methods, a Neural Network using a 1000-millisecond stream achieved the highest accuracy rate of 97.13 %, while a Random Forest using a 200-millisecond stream achieved the highest efficiency with an accuracy rate of 84.77 %. Furthermore, the study demonstrated that the computational resources required for this approach were minimal, making it ideal for use in XR applications. Overall, these findings represent a significant breakthrough in the field and have the potential to revolutionize the way XR applications are developed and implemented.